\documentclass[cameraready]{Interspeech}

\usepackage{comment}
\usepackage{booktabs}      
\usepackage{enumitem}      
\usepackage{threeparttable} 
\usepackage{multirow}

\title{Unified Gradient Projection: Language-Balanced Continual Learning for Multilingual Low-Resource ASR
\thanks{This work was supported by the National Natural Science Foundation of China under Grant No. 62276153.}
}

\author[orcid=0009-0008-9099-5334, equalcontribution]{Ziang}{Ren}

\author[orcid=0009-0001-1632-8428,equalcontribution]{Guodong}{Lin}
\author[orcid=0009-0002-6683-932X]{Yuchen}{Ai}
\author[orcid=0009-0003-9776-7303]{Kaize}{Tan}
\author[orcid=0000-0003-3841-1959, correspondingauthor]{Wei-Qiang}{Zhang}

\address{
    Department of Electronic Engineering, Tsinghua University, China
}

\email{ziangren65@gmail.com, wqzhang@tsinghua.edu.cn}

\keywords{Continual Learning,  Multilingual ASR, Low-Resource Speech, Gradient Projection}

\begin{document}

\maketitle

\begin{abstract}
    % 背景与痛点：
% 尽管 Whisper 等大规模预训练模型多语言识别能力卓越，但在微调适配低资源语种时，极易对历史知识产生灾难性遗忘 (Catastrophic Forgetting)。

% 现有局限（优化啰嗦问题）：
% 现有的持续学习方法存在两难困境：经验回放 (ER) 直接混合旧数据，但缺乏对参数更新方向的显式约束，稳定性不足；而 A-GEM 等梯度约束方法虽能缓解冲突，却需维护独立的任务基空间，在 Whisper 这种大规模架构中面临不可接受的显存扩展瓶颈。

% 本文方法（明确 UGP 与 ER 的区别与互补）：
% 为此，我们提出统一梯度投影机制 (Unified Gradient Projection, UGP)。UGP 同样利用少量回放数据，但其通过构建一个全局的“混合语言参考梯度”进行正交投影约束。它以极低的显存开销解决了参数更新时的梯度冲突，并与 ER 形成了**“梯度级约束”与“数据级混合”**的完美互补。

% 核心结论：
% 通过横跨 Small 到 Large-v3 的多尺度 (Scale-dependent) 分析，实验表明，UGP 在保持新语言极高可塑性的同时显著降低了遗忘率；特别是在 Large-v3 模型上，展现出近乎零遗忘的卓越稳定性。
Large-scale pretrained ASR models such as Whisper exhibit strong multilingual capabilities. However, fine-tuning on low-resource languages often causes catastrophic forgetting. Although continual learning mitigates this issue, existing methods struggle to regulate cross-task interference in multilingual settings, where dominant languages bias optimization.
We propose Unified Gradient Projection (UGP), which constrains parameter updates using reference gradients from language-balanced replay in a unified projection space. By equalizing per-language contributions in the projection, UGP reduces dominant-language bias and improves cross-lingual stability. We further show that combining gradient-level projection with data-level replay yields complementary gains in stability and plasticity.
Across diverse low-resource language groups and model scales, UGP enables effective adaptation while substantially mitigating forgetting. On Whisper-large-v3, it achieves near-zero average forgetting.
\end{abstract}

\section{Introduction}

The field of Automatic Speech Recognition (ASR) is undergoing a paradigm shift driven by large-scale foundation models\cite{yang2024large,babu2021xlsrselfsupervisedcrosslingualspeech}.
OpenAI's Whisper~\cite{radford2023robust}, for example, leverages massive weakly supervised data and a unified sequence-to-sequence architecture to achieve remarkable cross-lingual generalization within a single model.
Such scaling has significantly narrowed performance gaps across many languages. Nevertheless, performance on numerous low-resource languages remains sub-optimal~\cite{pratama2024analysis, liu2024exploration}, posing a fundamental challenge to the vision of truly universal speech recognition.

Adapting foundation models to under-represented languages through fine-tuning can substantially improve recognition accuracy~\cite{zhao2022improving}. However, sequential multilingual adaptation inevitably introduces the plasticity–stability dilemma~\cite{mermillod2013stability}: improving plasticity for newly encountered languages often compromises stability on previously learned ones.
As parameters shift to accommodate new linguistic patterns, representations essential to prior languages may be overwritten, leading to catastrophic forgetting~\cite{french1999catastrophic}. This tension becomes particularly pronounced in large multilingual ASR models~\cite{li2022massively}, where shared representations must support heterogeneous phonetic and linguistic structures.

% Continual learning (CL)~\cite{parisi2019continual} offers various strategies to balance the plasticity–stability trade-off, yet existing strategies remain imperfect in large-scale multilingual ASR models~\cite{vander2023using, libera2024clmasr}.
% Parameter-isolation techniques such as LoRA~\cite{hu2022lora} retain plasticity close to full fine-tuning while achieving limited stability through low-rank adaptation, but they do not explicitly constrain cross-task interference and may reduce direct parameter sharing across language~\cite{ban2025rethinkingparametersharingllm}.
% Regularization-based methods like EWC~\cite{kirkpatrick2017overcoming, qian2024learn} promote stability by penalizing changes to important parameters, though they rely on approximate importance estimation and incur substantial overhead in large models.
% Replay-based paradigms, particularly Experience Replay (ER)~\cite{rolnick2019experience}, mitigate forgetting by mixing historical and current data during training, yet they lack explicit control over gradient interactions and may struggle to maintain stability under severe gradient conflicts~\cite{dellalibera2024clmasr}.

Continual learning (CL)~\cite{parisi2019continual} offers various strategies
for the plasticity--stability trade-off, yet existing approaches remain
imperfect in large-scale multilingual ASR~\cite{vander2023using, libera2024clmasr}.
Parameter-isolation methods such as LoRA~\cite{hu2022lora} retain plasticity
but do not explicitly constrain cross-task interference and may reduce parameter
sharing across languages~\cite{ban2025rethinkingparametersharingllm};
regularization-based methods like EWC~\cite{kirkpatrick2017overcoming, qian2024learn}
promote stability but rely on approximate importance estimation and incur
substantial overhead at model scale.
Replay-based paradigms, particularly Experience Replay
(ER)~\cite{rolnick2019experience}, mitigate forgetting through data rehearsal,
yet lack explicit gradient control and may struggle under severe gradient
conflicts~\cite{libera2024clmasr}.

Among these paradigms, gradient-based continual learning regulates interference by constraining update directions. Methods such as GEM~\cite{lopez2017gradient} and its scalable variant A-GEM~\cite{chaudhry2018efficient} preserve prior knowledge by projecting gradients using reference information from previous tasks, with A-GEM approximating multiple constraints through a single reference gradient for improved scalability.
However, multilingual ASR inherently involves imbalanced language distributions in both pretraining data and replay buffers. Under such heterogeneity, reference gradient estimation can be biased toward dominant languages, skewing update directions and resulting in uneven stability~\cite{yu2020gradient}, where low-resource tasks remain more susceptible to forgetting~\cite{zhang2024proactivegradientconflictmitigation}. As model capacity increases, richer representations further amplify cross-lingual gradient interactions, making balanced gradient regulation increasingly critical.

Motivated by this observation, we propose Unified Gradient Projection (UGP), a unified continual learning framework for multilingual ASR that integrates language-balanced gradient regulation with ER.
At the gradient level, UGP projects parameter updates into relatively independent subspaces, effectively decoupling cross-lingual knowledge to mitigate inter-language interference~\cite{yu2020gradient}. At the data level, UGP synergizes with ER, which prevents catastrophic forgetting while simultaneously acting as joint multilingual training to promote positive knowledge transfer for low-resource target language~\cite{pekarek2023replay}.

Our main contributions are summarized as follows:

\begin{enumerate}
\item \textbf{Language-Balanced Gradient Regulation}:

We introduce a novel language-balanced gradient regulation strategy. By estimating a language-balanced holistic reference gradient, this strategy reshapes the optimization geometry toward less antagonistic cross-lingual gradient interactions.

\item \textbf{Unified Gradient Projection (UGP)}:

We propose Unified Gradient Projection (UGP), a unified framework that embeds Experience Replay within the language-balanced gradient regulation mechanism. Projection regulates destructive update directions at the optimization level, while ER mitigates harmful drift on previously learned languages and enables data-level joint training to encourage positive cross-lingual transfer. This coordinated design improves the stability–plasticity trade-off for low-resource multilingual adaptation.

\item \textbf{Scale-Dependent \& Low-Resource Analysis}:

Through systematic empirical studies from Whisper-small to large-v3, we reveal that larger-capacity models empower UGP to achieve near-zero average forgetting. Furthermore, we demonstrate UGP's exceptional structural stability even under extreme data scarcity (down to 5 hours), suggesting its robustness under severe data scarcity.

\end{enumerate}

\section{Method}
% 3.1 Foundational Fine-tuning Strategy (基础微调策略)保留原样：保留冻结编码器（Frozen Encoder）和数据增强（SpecAugment）的论述。这为后续 UGP 为什么能在一个鲁棒的子空间中工作奠定了基础。
% 3.2 Unified Gradient Projection (UGP 核心机制)痛点切入 (The Bottleneck)：首先点名批评 GPM 和 A-GEM。指出它们在应对大模型时，最大的败笔在于需要存储正交子空间矩阵 (Subspace Storage) 或针对每个旧任务计算独立梯度 (Per-task Constraint)，导致显存随任务数线性爆炸 $\mathcal{O}(N)$。核心定义 (The Mechanism)：给出 UGP 的正交投影公式。三大亮点提炼 (Key Innovations)：使用项目符号（Bullet points / itemize）将你提出的三个亮点单独列出，加粗强调。这是论文写作的“吸睛”技巧，Reviewer 扫读时一眼就能看到。No Subspace Storage (免子空间存储)：强调显存开销仅为 $\mathcal{O}(1)$。Task-Agnostic Constraint (摒弃 Per-task 约束)：说明不需要人为划分历史任务边界。Dynamic Holistic Reference (动态全局约束)：解释混合 batch 如何提供全局的防遗忘护栏。
% 3.3 Synergistic Integration with Experience Replay (与经验回放的协同)简短补充：既然不写 LaBER，只需用一段话简述 UGP 作为一个“梯度级硬约束”，可以与任何标准的数据级 ER（Experience Replay）无缝结合，实现“表征修复”与“方向纠正”的双管齐下。
% In this section, we introduce LaBER and UGP, two novel fine-tuning strategies that leverage a frozen encoder.
In this section, we present the proposed Unified Gradient Projection (UGP), a unified continual learning framework for multilingual ASR that integrates gradient-level interference regulation with data-level replay.

\subsection{Foundational Fine-tuning Overview}
% Our foundational fine-tuning strategy freezes the encoder to preserve its powerful, pre-trained feature extraction capabilities, a technique known to effectively mitigate catastrophic forgetting \cite{liu2024exploration, takashima2022updating}. Consequently, updates are exclusively applied to the decoder, positional encodings, and token embeddings. To enhance model robustness and data efficiency, our strategy also incorporates several data augmentation methods, including stochastic time stretch, volume disturbance, Gaussian noise, and SpecAugment \cite{park2019specaugment}.
To maintain the pre-trained feature extraction capability of the model, we adopt a foundational fine-tuning strategy for medium and large models, where the encoder is frozen and only the decoder and embeddings are updated. For Whisper-small, full-parameter fine-tuning is applied to avoid capacity under-utilization.
% Specific experimental settings, including data augmentation techniques, are detailed in Section~\ref{sec:experiments}.

\subsection{Gradient-Level Language-Balanced Projection}
Catastrophic forgetting arises from conflicting gradients between the current target language and previously learned languages. 
To address this, UGP performs gradient-level interference regulation through a single, dynamically constructed reference gradient.

Let $g_{\text{cur}}$ denote the gradient of the current training batch, and $g_{\text{ref}}$ denote a reference gradient computed from a language-balanced mixture of replay samples representing past knowledge. 
Compared with A-GEM, which approximates constraints from randomly sampled historical data, this balanced reference ensures that each language contributes equally when estimating $g_{\text{ref}}$, mitigating dominant-language bias.
To construct a language-balanced reference gradient, UGP uniformly samples the same number of replay samples from each historical language at every training step.
Formally, assume the replay buffer contains $K$ languages with sample sets $B_1, B_2, \dots, B_K$. 
For each language $i$, we draw $n$ samples uniformly at random:
\begin{equation}
    \tilde{B}_i \sim \text{UniformSample}(B_i, n), \quad i=1,\dots,K.
\end{equation}

The reference gradient is then computed as:

\begin{equation}
    \label{eq:g_ref}
    g_{\text{ref}} = \frac{1}{K n} \sum_{i=1}^{K} \sum_{x \in \tilde{B}_i} \nabla_\theta \mathcal{L}(x).
\end{equation}

Interference is detected by the inner product $\langle g_{\text{cur}}, g_{\text{ref}} \rangle$. If the gradients conflict (i.e., the angle is obtuse), the current gradient is projected onto the orthogonal complement of $g_{\text{ref}}$:
\begin{equation}
    \label{eq:ugp}
    g_{\text{final}} =
    \begin{cases}
        g_{\text{cur}} & \text{if } \langle g_{\text{cur}}, g_{\text{ref}} \rangle \geq 0, \\[8pt]
        g_{\text{cur}} - \dfrac{\langle g_{\text{cur}}, g_{\text{ref}} \rangle}{\Vert g_{\text{ref}} \Vert^2 + \epsilon} \cdot g_{\text{ref}} & \text{if } \langle g_{\text{cur}}, g_{\text{ref}} \rangle < 0,
    \end{cases}
\end{equation}
where $\epsilon$ is a small constant for numerical stability.

\begin{figure}[htbp]
    \centering
    \vspace{-0.5em}
    % Replace with your UGP diagram
    \includegraphics[trim=1cm 10cm 1cm 0.5cm, clip, width=0.9\columnwidth]{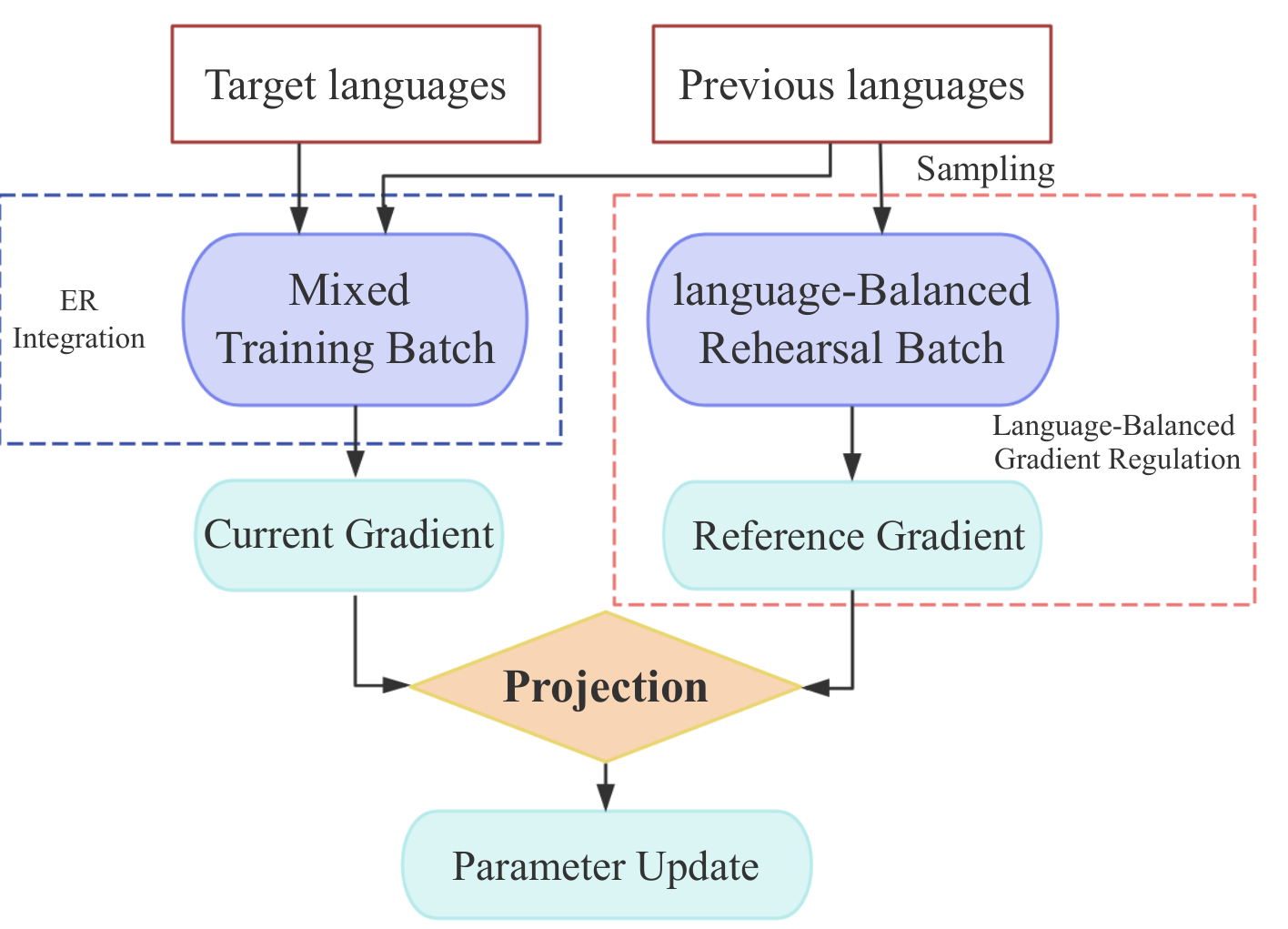}
    \caption{Workflow of the proposed Unified Gradient Projection (UGP), highlighting the dynamic holistic constraint.}
    \label{fig:ugp}
    \vspace{-1.28em}
\end{figure}

\subsection{Integration with Experience Replay}
UGP is designed to integrate seamlessly with data-level ER. 
For each update, a mixed batch is constructed by uniformly sampling historical data from all languages in the replay buffer, ensuring balanced coverage across tasks. 
ER provides data-driven representation correction via standard loss optimization, while gradient-level language-balanced projection regulates gradient directions to mitigate cross-task interference.

From an optimization perspective, this combined mechanism can be interpreted as a two-level regulation process. 
ER modifies the empirical training objective by augmenting the current-language loss with replayed samples, yielding a mixed objective:

\begin{equation}
    \mathcal{L}_{\text{total}} 
    = \mathcal{L}_{\text{cur}} 
    + \lambda \mathcal{L}_{\text{replay}},
\end{equation}
where $\lambda$ controls the relative contribution of historical data. In our setting, $\lambda = 1$.
This objective-level mixing anchors parameter updates to previously learned tasks.

Gradient-Level Language-Balanced Projection further regulates the resulting gradients by removing components that conflict with balanced historical knowledge. 
Consequently, ER shapes the optimization landscape through data rehearsal, while Gradient-Level Language-Balanced Projection governs the trajectory taken within that landscape.

Together, these two mechanisms form the complete Unified Gradient Projection (UGP) framework, which:

\begin{itemize}
    \item Reduces catastrophic forgetting by preserving historical knowledge.
    \item Encourages positive transfer across languages by promoting optimization directions beneficial to multiple tasks.
    \item Maintains strong adaptation to the current target language without diluting gradients or requiring additional memory overhead.
\end{itemize}

\begin{table*}[t]
    \centering
    \footnotesize % 缩小一号字体
    \setlength{\tabcolsep}{4pt} % 缩小列间距
    \caption{We report WER on the target languages (TWER), previous languages (RWER), average (AWER), and the degree of catastrophic forgetting (FWER) across three Whisper scales on the Southeast Asian core set. Target: Malay (10.47h), Indonesian (10.24h), Filipino (9.68h), Javanese (11.33h), Māori (20.48h). Replay: Thai (2h), Vietnamese (2h), English (2h), French (2h).}
    \label{tab:main_results}
    \begin{tabular}{l @{\hskip 6pt}cccc @{\hskip 6pt}cccc @{\hskip 6pt}cccc}
        \toprule
        & \multicolumn{4}{c}{\textbf{whisper-small}} & \multicolumn{4}{c}{\textbf{whisper-medium}} & \multicolumn{4}{c}{\textbf{whisper-large-v3}} \\
        \cmidrule(lr){2-5} \cmidrule(lr){6-9} \cmidrule(lr){10-13}
        \textbf{Method} & \textbf{TWER}  & \textbf{RWER}  & \textbf{AWER} $\downarrow$ & \textbf{FWER} $\downarrow$ & \textbf{TWER}  & \textbf{RWER}  & \textbf{AWER} $\downarrow$ & \textbf{FWER} $\downarrow$ & \textbf{TWER}  & \textbf{RWER}  & \textbf{AWER} $\downarrow$ & \textbf{FWER} $\downarrow$ \\
        \midrule
        Vanilla Whisper & 46.11 & 16.42 & 31.26 & -- & 46.76 & 10.50 & 28.63 & -- & 26.75 & 6.65 & 16.70 & -- \\
        \midrule
        FT & \textbf{20.80} & 70.83 & 45.82 & 54.42 & 18.61 & 99.80 & 59.20 & 89.80 & \textbf{11.40} & 15.63 & 13.51 & 8.98 \\
        Standard ER & \underline{22.62} & \underline{36.14} & \underline{29.38} & \underline{19.72} & 17.06 & \textbf{18.82} & \textbf{17.94} & \textbf{8.82} & \underline{12.86} & \underline{10.75} & \underline{11.81} & \underline{4.11} \\
        Standard A-GEM & 23.70 & 73.56 & 48.63 & 57.14 & \textbf{15.86} & 25.51 & 20.68 & 15.51 & 22.20 & 16.43 & 19.32 & 9.78 \\
        \midrule
        \textbf{UGP (Ours)} & 23.26 & \textbf{28.36} & \textbf{25.81} & \textbf{11.94} & \underline{16.50} & \underline{21.05} & \underline{18.78} & \underline{10.05} & 12.91 & \textbf{6.68} & \textbf{9.80} & \textbf{0.04} \\
        \bottomrule
    \end{tabular}
\end{table*}

\section{Experiments}
\label{sec:experiments}
To evaluate our continual learning strategies, we conducted experiments across three model scales: Whisper-small (244M), medium (769M), and large-v3 (1550M).

\subsection{Datasets and Scenarios}
To comprehensively assess scalability and robustness, we designed two distinct experimental scenarios primarily utilizing the FLEURS dataset \cite{conneau2023fleurs}.

\textbf{Core Evaluation Set:} For the main cross-scale evaluation (Small to Large-v3), we constructed a highly interfering set comprising five target languages (Malay, Indonesian, Filipino, Javanese, Māori) and four previous languages (Thai, Vietnamese, English, French). 

\textbf{Extended \& Data Scaling Set:} To further validate generalizability and investigate the impact of extreme data scarcity, we introduced an extended language set evaluated on the Whisper-small model. The target group consists of Swahili, Persian, and Arabic, representing distinct language families, while the previous group includes English, French, and Russian. Furthermore, to conduct a rigorous data-scaling analysis, we mixed the FLEURS and CommonVoice\cite{commonvoice:2020} datasets to create precisely controlled subsets of 50h, 30h, 10h, and 5h per language.

\subsection{Evaluation Metrics}
We evaluate the performance using Word Error Rate (WER), noting that Character Error Rate (CER) is reported for Thai due to its lack of word boundaries. We report four key metrics. Target WER (TWER) and Retention WER (RWER) denote the average WER across the target and rehearsal languages, reflecting the model's plasticity and stability. Average WER (AWER) represents the group-balanced overall performance between target and rehearsal languages, defined as:

% \begin{equation}
% \label{eq:awer_metric}
% \text{AWER} = \frac{1}{|\mathcal{L}_{\text{all}}|} \sum_{l \in \mathcal{L}_{\text{all}}} \text{WER}_{l}
% \end{equation}
\begin{equation}
\mathrm{AWER} = \frac{1}{2}(\mathrm{TWER} + \mathrm{RWER}).
\end{equation}
To precisely quantify catastrophic forgetting on previous languages $\mathcal{L}_{\text{prev}}$, we use Forgetting WER (FWER)~\cite{libera2024clmasr}, formulated as:
\begin{equation}
\label{eq:fwer_metric}
\text{FWER} = \frac{1}{|\mathcal{L}_{\text{prev}}|} \sum_{l \in \mathcal{L}_{\text{prev}}} ( \text{WER}_{l}^{\text{end}} - \text{WER}_{l}^{\text{min}} )
\end{equation}
where $\text{WER}_{l}^{\text{end}}$ is the final WER after all training, and $\text{WER}_{l}^{\text{min}}$ is the historical best WER for language $l$.

\subsection{Baselines}
We compare our proposed UGP and its variant without Experience Replay (UGP w/o ER) against four established baselines: Vanilla Whisper, Vanilla FT (full-parameter fine-tuning on target languages only), Standard ER, and Standard A-GEM.

% \subsection{Implementation Details}
% Models were trained on NVIDIA A40 GPUs using the AdamW optimizer \cite{loshchilov2017decoupled} with early stopping (patience=3). To ensure methodological consistency during the data-scaling experiments (50h to 5h), early stopping was uniformly applied across all subsets to rigorously evaluate the genuine forgetting rate under optimal target plasticity. For the medium and large-v3 models, we utilized a frozen encoder. For the small model, to prevent over-constraining its limited capacity, we employed full-parameter fine-tuning. Data was doubled via waveform augmentation, and SpecAugment~\cite{park2019specaugment} was strictly applied to all target samples to enhance gradient variance.
\subsection{Implementation Details}
Models were trained on NVIDIA A40 GPUs using the AdamW
optimizer~\cite{loshchilov2017decoupled} with early stopping (patience=3),
applied uniformly across all data-scaling subsets to evaluate the genuine
forgetting rate at optimal plasticity.
The encoder was frozen for medium and large-v3 models;
full-parameter fine-tuning was applied to the small model
to avoid capacity under-utilization.
For UGP, $n{=}4$ utterances are drawn per historical language per step,
with each per-language reference pool capped at 2{,}000 utterances;
the ER buffer contains 2{,}000 utterances distributed uniformly
across prior languages.
Data was doubled via waveform augmentation, and
SpecAugment~\cite{park2019specaugment} was applied to all target samples.

\section{Results}
\label{sec:results}

\begin{table*}[htbp]
\centering
\small
\setlength{\tabcolsep}{6pt}
\caption{Data-scaling (50h to 5h) and language extension evaluation on low-resource targets (Swahili, Persian, Arabic) against replay languages (English, French, Russian) using Whisper-small. Evaluated on the Common Voice testset, reporting Average WER (AWER) and Forgetting WER (FWER).}
\label{tab:todo_scaling}
\begin{tabular}{@{}l cc cc cc cc@{}}
\toprule
& \multicolumn{2}{c}{\textbf{50 Hours}} & \multicolumn{2}{c}{\textbf{30 Hours}} & \multicolumn{2}{c}{\textbf{10 Hours}} & \multicolumn{2}{c}{\textbf{5 Hours}} \\
\cmidrule(lr){2-3} \cmidrule(lr){4-5} \cmidrule(lr){6-7} \cmidrule(lr){8-9}
\textbf{Method} & \textbf{AWER} & \textbf{FWER} & \textbf{AWER} & \textbf{FWER} & \textbf{AWER} & \textbf{FWER} & \textbf{AWER} & \textbf{FWER} \\
\midrule
Full FT         & 79.59 & 112.11 & 83.59 & 112.39 & 60.96 & 63.86 & 50.24 & 35.47 \\
Standard ER     & 34.12 & 27.88 & 36.69 & 20.75 & 41.39 & 14.07 & 45.11 & 10.73 \\
Standard A-GEM  & 36.54 & 27.43 & 38.91 & 22.41 & 41.50 & 15.15 & 44.49 & 12.70 \\
\midrule
UGP (wo ER) & 35.93 & 26.32 & 38.70 & 20.02 & \textbf{35.73} & 12.76 & \textbf{37.89} & 11.28 \\
\textbf{UGP(ours)}          & \textbf{31.47} & \textbf{15.62}     & \textbf{35.66} & \textbf{12.09} & 38.48 & \textbf{9.22}  & 43.40 & \textbf{8.17} \\
\bottomrule
\end{tabular}
\end{table*}

\subsection{Overall Performance Across Scales}

As shown in Table~\ref{tab:main_results}, UGP achieves a strong stability--plasticity trade-off across scales, obtaining the best overall results on Whisper-small and Whisper-large-v3 while remaining competitive on Whisper-medium.

Specifically, standard full-parameter fine-tuning (FT) suffers from severe catastrophic forgetting, reaching an FWER of 89.80\% on the medium model. Standard ER offers limited mitigation, and while Standard A-GEM protects past knowledge, it severely limits target language plasticity and inflates the TWER. UGP successfully overcomes this trade-off across all scales. Notably, its advantage increases with model capacity. On Whisper-large-v3, UGP exploits the model's vast representational flexibility to achieve near-zero average forgetting with an FWER of just 0.04\%. Simultaneously, it maintains highly competitive target plasticity with a TWER of 12.91\%, yielding an exceptional overall AWER of 9.80\%.

\begin{figure}[htbp]
    \centering
    \includegraphics[width=\columnwidth]{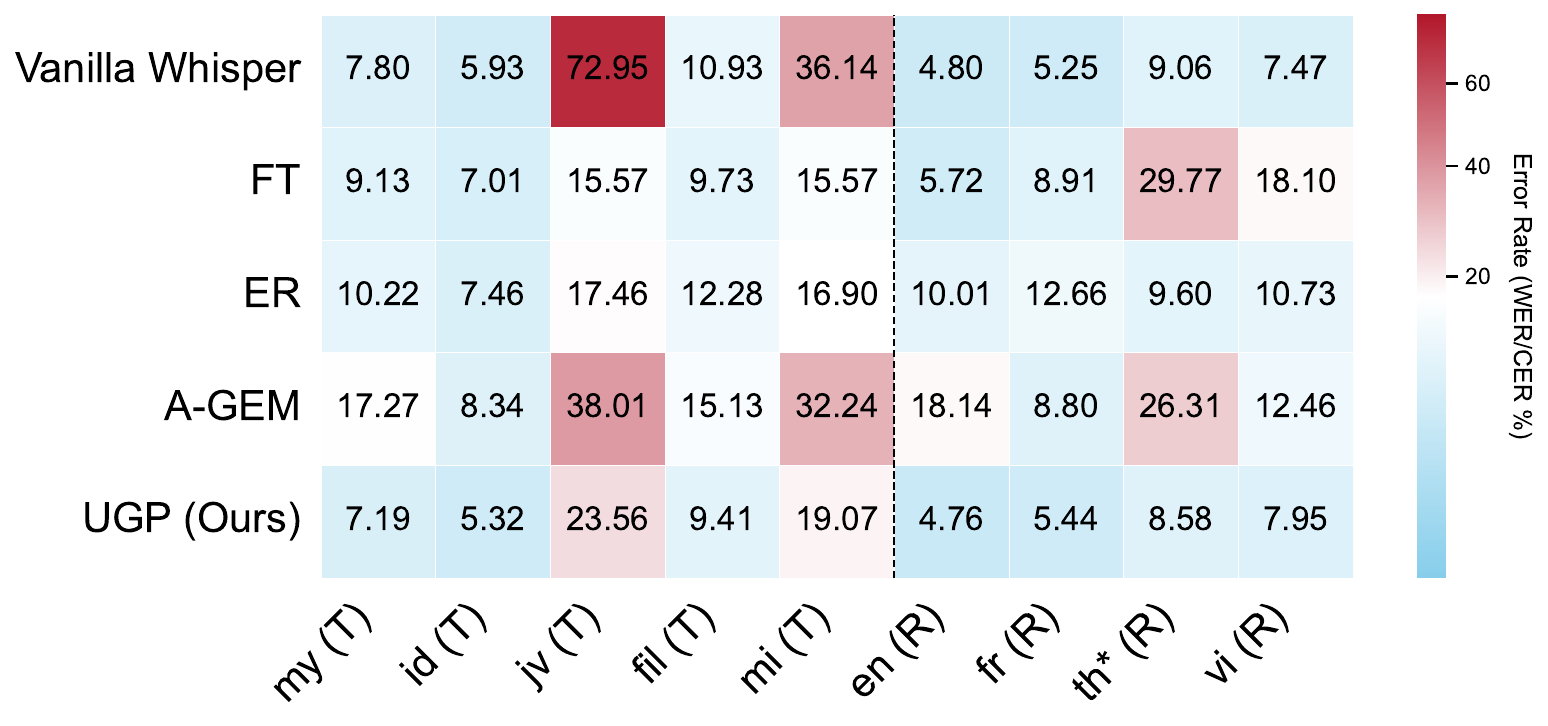} 
    \caption{Per-language Word Error Rate (\%) comparison of different methods on Whisper-large-v3. Target newly-learned languages and previously learned languages are denoted by (T) and (R), respectively. The asterisk (*) indicates that CER is reported for Thai (th).}
    \label{fig:heatmap_whisperlarge}
\end{figure}

This optimal equilibrium is clearly visualized in the Whisper-large-v3 per-language heatmap in Figure~\ref{fig:heatmap_whisperlarge}. Unlike standard FT, which causes catastrophic degradation in historical tasks—highlighted by sharp error spikes like the prominent red region for Thai—UGP effectively alleviates this interference. It preserves baseline performance across all historical languages while significantly reducing error rates on target languages. This demonstrates UGP's ability to safely adapt foundation models to new domains without overriding critical historical representations.

\subsection{Ablation Study}

\begin{table}[htbp]
\centering
\small
\setlength{\tabcolsep}{4pt}
\caption{Ablation study on Whisper-large-v3.}
\label{tab:ablation_study}
\begin{tabular}{@{}l c c@{}}
\toprule
\textbf{Configuration} & \textbf{TWER (\%)} & \textbf{FWER (\%)} \\
\midrule
Vanilla FT & 13.48 & 13.79 \\
\midrule
FT + Augment + Freeze Enc. (\textbf{Base}) & 19.04 & 2.77 \\
\midrule
Base + Standard ER & 12.86 & 4.11 \\
Base + Standard A-GEM & 22.20 & 9.78 \\
\midrule
Base + Unified Proj. (w/o ER) & 15.80 & 4.20 \\
Base + Unified Proj. + ER (\textbf{UGP}) & \textbf{12.91} & \textbf{0.04} \\
\bottomrule
\end{tabular}
\vspace{-1.0em}
\end{table}

Table~\ref{tab:ablation_study} details our component-wise ablation study on Whisper-large-v3. Compared with vanilla fine-tuning in this ablation setting, the baseline with a frozen encoder and data augmentation reduces forgetting but limits target plasticity. Integrating ER restores this plasticity because the replay mechanism increases the total training data volume and drives positive cross-lingual transfer through multilingual joint training\cite{pekarek2023replay}. However, its unconstrained optimization slightly increases forgetting. Meanwhile, standard A-GEM preserves past knowledge but severely stifles target adaptation. Replacing it with our unified projection resolves this bottleneck, suggesting that a holistic, language-balanced reference gradient generalizes far better. Ultimately, the complete UGP framework synergizes these approaches. It maintains strong target plasticity while achieving a near-zero forgetting rate of 0.04\%, demonstrating that data-level representation alignment and gradient-level interference correction are highly complementary.

\subsection{Mechanism Analysis: Gradient Decoupling and Orthogonalization}

% 插入您原有的图表代码
% \begin{figure}[htbp]
%     \centering
%     \includegraphics[width=\columnwidth]{figure/pairwise_cosine_heatmap.pdf} 
%     \caption{Pairwise gradient cosine similarity between target and historical languages during the adaptation of Whisper-large-v3. Our proposed UGP (left) successfully stabilizes updates near orthogonality, while standard Full Fine-Tuning (right) suffers from severe gradient conflicts.}
%     \label{fig:heatmap_cosine}
% \end{figure}

\begin{figure}[htbp]
    \centering
    \includegraphics[width=\columnwidth]{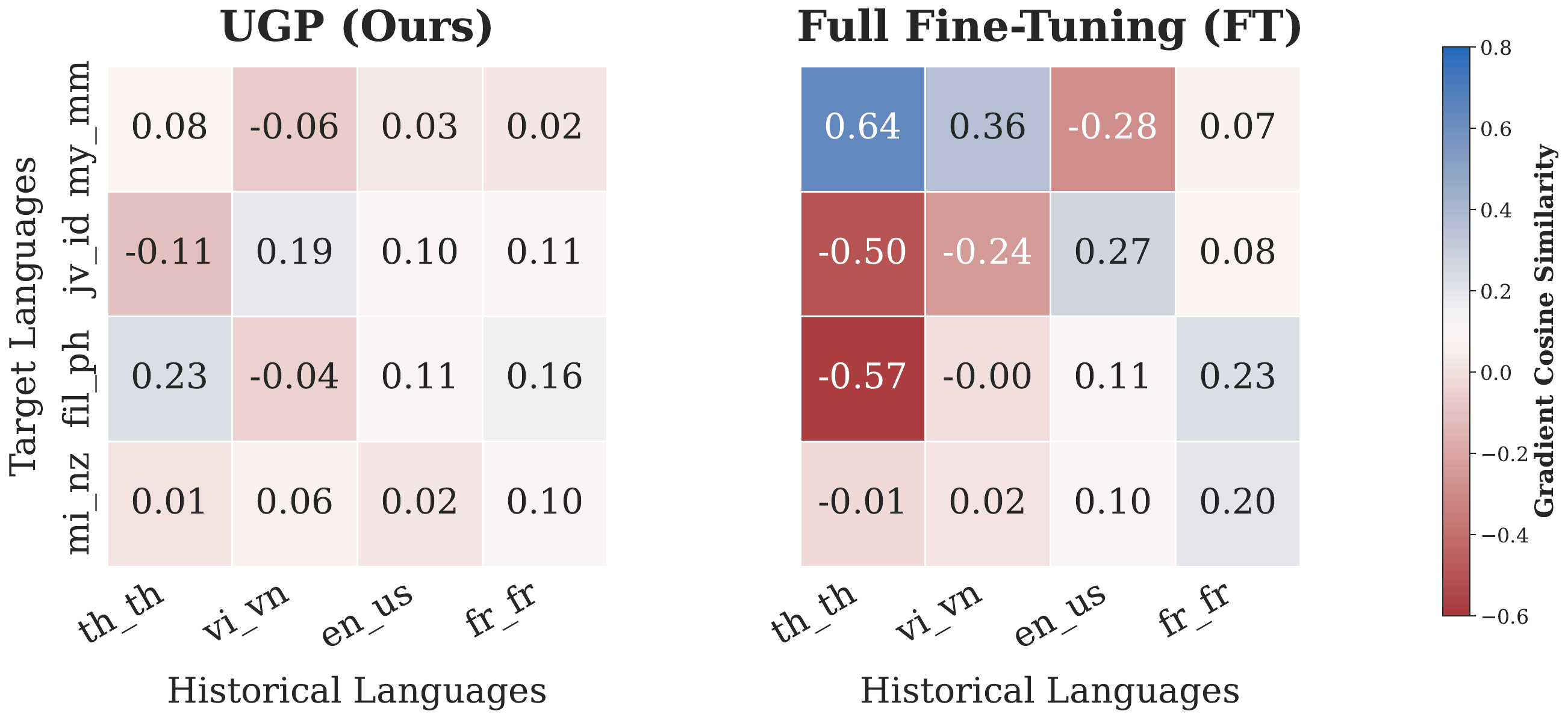} 
    \caption{Pairwise gradient cosine similarity between a representative subset of target and historical languages during the adaptation of Whisper-large-v3. Our proposed UGP (left) stabilizes the analyzed updates near orthogonality, while standard Full Fine-Tuning (right) exhibits stronger gradient conflicts in this subset.}
    \label{fig:heatmap_cosine}
\end{figure}

% 对图表的精炼解释
% To further investigate the underlying mechanism of UGP, we conducted 50 inference sampling runs on the Whisper-large-v3 model adapted via UGP and standard Full Fine-Tuning, analyzing the pairwise gradient cosine similarity between target and historical languages. As illustrated in Figure~\ref{fig:heatmap_cosine}, standard FT exhibits severe gradient conflicts, characterized by significant negative cosine values, which directly drive catastrophic forgetting. In stark contrast, UGP successfully constrains the gradient update directions across different languages to near-orthogonality. This structural stabilization proves that UGP does more than merely slow down forgetting; it effectively pulls language representations into independent, mutually non-interfering orthogonal subspaces, achieving superior knowledge decoupling and mitigating cross-lingual interference at the parameter level.
% To examine how UGP affects the learned parameter geometry, we analyze the fully trained Whisper-large-v3 models obtained via UGP and standard Full Fine-Tuning (FT). After convergence, we compute pairwise gradient cosine similarities between target and historical languages. As illustrated in Figure~\ref{fig:heatmap_cosine}, the FT-trained model exhibits persistent negative cosine values, indicating that its converged parameter configuration remains prone to cross-lingual gradient conflicts. In contrast, the UGP-trained model yields gradient interactions concentrated near orthogonality.
To examine how UGP affects the learned parameter geometry, we analyze the fully trained Whisper-large-v3 models obtained via UGP and standard Full Fine-Tuning (FT). After convergence, we compute pairwise gradient cosine similarities between a representative subset of target and historical languages. As illustrated in Figure~\ref{fig:heatmap_cosine}, the FT-trained model exhibits persistent negative cosine values in this subset, indicating that its converged parameter configuration remains prone to cross-lingual gradient conflicts. In contrast, the UGP-trained model yields gradient interactions concentrated near orthogonality.

This suggests that UGP does not merely constrain individual updates during training; rather, it shapes the optimization landscape toward a parameter state in which language-specific gradients are structurally less antagonistic, thereby reducing destructive interference at convergence.

\subsection{Language Extension and Extreme Data Scaling}

Due to computational constraints, data-scaling analyses (50h to 5h) were evaluated on Whisper-small. While UGP ensures robust knowledge preservation under extreme scarcity, Table~\ref{tab:todo_scaling} reveals that increasing data volume significantly boosts ER's positive transfer to target languages. Crucially, UGP effectively inherits this amplified plasticity while suppressing the severe catastrophic forgetting seen in Full FT. Consequently, UGP consistently minimizes catastrophic forgetting across all data regimes, yielding robust overall performance (AWER) even under extreme data scarcity.

\section{Conclusion}

To mitigate catastrophic forgetting in multilingual ASR, we propose Unified Gradient Projection (UGP). By combining language-balanced projection with Experience Replay, UGP orthogonalizes conflicting updates to achieve near-zero forgetting on Whisper-large-v3, providing an efficient solution for universal speech recognition.

\section{Generative AI Use Disclosure}

Generative AI tools were used only to assist with language polishing and to support the analysis and interpretation of related materials. All research ideas, experimental design, implementation, experiments, data analysis, results, and conclusions were independently completed and verified by the authors.

\bibliographystyle{IEEEtran}
\bibliography{mybib}

\end{document}